\documentclass[runningheads]{llncs}
\usepackage{tikz}
\usepackage{comment}
\usepackage{amsmath,amssymb} 
\usepackage{color}

\usepackage[accsupp]{axessibility}  

\usepackage{microtype} 

\usepackage{adjustbox}
\usepackage{listings}
\usepackage{pythonhighlight}
\usepackage{float}
\usepackage{tablefootnote}
\usepackage{tabularx}
\usepackage{booktabs}
\usepackage{graphicx}
\usepackage{wrapfig}

\newcommand{\etal}{\textit{et al.}}

\newcommand{\TW}{$\mathtt{300W}$}
\newcommand{\COFW}{$\mathtt{COFW}$}

\newcommand{\WFLW}{$\mathtt{WFLW}$}
\newcommand{\LAPA}{$\mathtt{LaPa}$}
\newcommand{\CaricatureFace}{$\mathtt{CariFace}$}
\newcommand{\AnimalWeb}{$\mathtt{AnimWeb}$}
\newcommand{\ArtisticFaces}{$\mathtt{ArtFace}$}
\newcommand{\PARE}{$\mathtt{PARE}$}
\newcommand{\MDMDBase}{MDMD Base}
\newcommand{\MDMD}{MDMD w/LaPa}
\usepackage[pagebackref=true,breaklinks=true,colorlinks,bookmarks=false]{hyperref}
\usepackage{multirow}
\usepackage{listings}
\usepackage[shortlabels]{enumitem}
\usepackage{hyperref}
\usepackage{adjustbox}
\usepackage[normalem]{ulem}

\begin{document}

\pagestyle{headings}
\mainmatter
\def\ECCVSubNumber{3780}

\title{Multi-Domain Multi-Definition Landmark Localization for Small Datasets}
\titlerunning{MDMD Landmark Localization}
\author{David Ferman\inst{1,2} \and
Gaurav Bharaj\inst{1}}
\authorrunning{D. Ferman and G. Bharaj}
\institute{AI Foundation, USA \and UT Austin, USA\\
\email{davidcferman@gmail.com}}
\maketitle

\begin{abstract}
We present a novel method for multi image domain and multi-landmark definition learning for small dataset facial localization. Training a small dataset alongside a large(r) dataset helps with robust learning for the former, and provides a universal mechanism for facial landmark localization for new and/or smaller standard datasets. To this end, we propose a Vision Transformer encoder with a novel decoder with a definition agnostic shared landmark semantic group structured prior, that is learnt, as we train on more than one dataset concurrently. Due to our novel definition agnostic group prior the datasets may vary in landmark definitions and domains. During the decoder stage we use cross- and self-attention, whose output is later fed into domain/definition specific heads that minimize a Laplacian-log-likelihood loss. We achieve state-of-the-art performance on standard landmark localization datasets such as \COFW~and \WFLW, when trained with a bigger dataset. We also show state-of-the-art performance on several varied image domain small datasets for animals, caricatures, and facial portrait paintings. Further, we contribute a small dataset (150 images) of pareidolias to show efficacy of our method. Finally, we provide several analysis and ablation studies to justify our claims.
\end{abstract}

\begin{figure}[t]
\center
  \includegraphics[width=\textwidth]{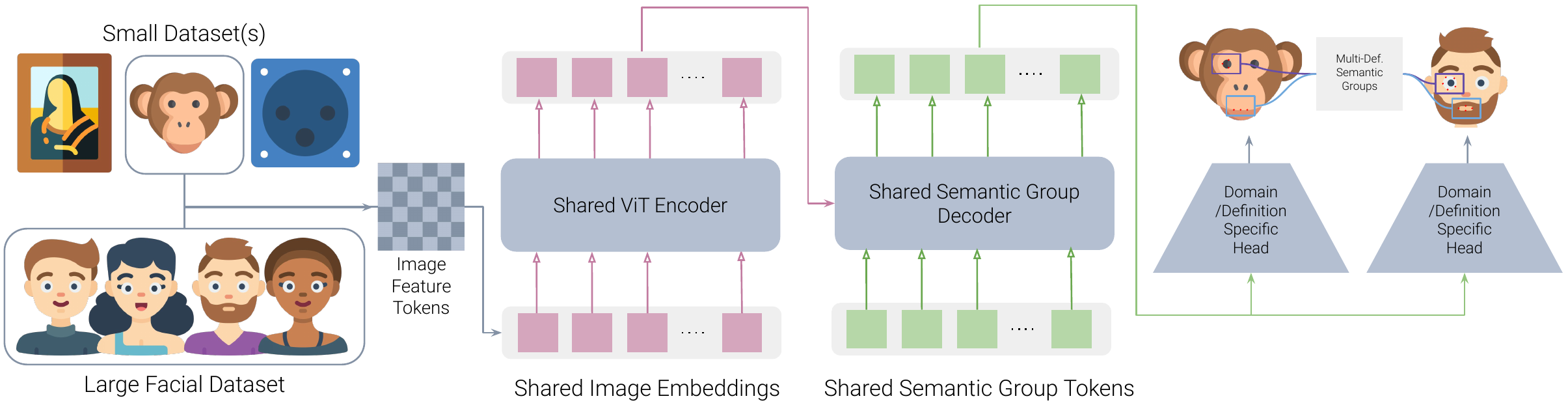}
  \caption{System overview (left to right): Our method takes a small dataset with a landmark definition, and a larger facial dataset with a different landmark definition, and relies on common semantic group definitions to learn for both dataset concurrently.}
  \label{fig:overview}
\end{figure}
\section{Introduction}
\label{sec:introduction}
With the rising need for novel AR/VR, telepresence, character animation filter applications (e.g., adding props and effects in live video streams of humans, pets, etc.), arises the need for facial localization for multiple image domains. While, supervised landmark localization has made great strides for the \emph{in-the-wild} human faces domain, it is often hard to create such datasets for new image domains -- animals Khan~\etal~\cite{Khan2020AnimalWebAL}, art~\cite{Wei2019VRFA}, cartoons, and more recently, pareidolias Song~\etal~\cite{song2021everything} that abstractly resemble human faces, Wardle~\etal~\cite{wardle2022illusory}. Building a dataset for supervised learning of landmarks is hard due to the cumbersome hand-labeling process, where, hand-labels can lead to noisy and inconsistent landmarks~\cite{Dong2018SupervisionbyRegistrationAU}, and is often very time consuming for new domains~\footnote{Labeling a landmark dataset for animal faces can take up to 6,833 hours~\cite{Khan2020AnimalWebAL}}.

\begin{figure}[t]
\centering
  \includegraphics[width=\textwidth]{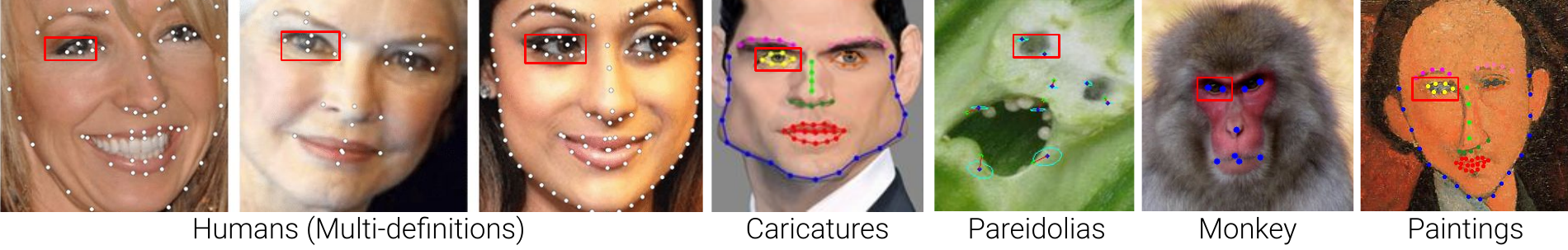}
  \caption{Our method works for multiple domains and multiple definitions of facial landmark localization problems. The {\color{red}red box} represents a \emph{landmark semantic group} shared across different domains and landmark definitions.}
  \label{fig:teaser}
\end{figure}

Due to varied new domains, Figure~\ref{fig:teaser}, and subsequent specific applications, there's no preset definition for facial landmarks. For example, a landmark definition set that works for humans faces may not work for animal faces and vice-versa and thus makes cross-domain learning infeasible. Additionally, within human face localization problems, different datasets have different definitions of landmarks, see Figure~\ref{fig:teaser} (Humans), and certain applications can require unique landmark definitions, e.g., landmarks which correspond to mesh vertices, Wei~\etal~\cite{Wei2019VRFA}. The landmark datasets necessitated by a particular new application are either small or non-existent. As a result, novel applications that need localization for new image domains and/or definitions becomes infeasible. Image domain localization problems have been previously approached with domain transfer methods. For example, Yaniv~\etal~\cite{Yaniv2019TheFO} use domain transfer to approach learning for facial portrait artwork, Wei~\etal~\cite{Wei2019VRFA} learn landmark correspondences as an auxiliary aspect of mesh fitting. Such methods need a specialized larger dataset and/or landmark definitions from a previous dataset, which might be sub-optimal for the candidate domain. In this work, we create a method that learns robust landmarks for new domains for which small datasets may exist, or for which a small set of labeled images can be obtained, inexpensively, while being landmark definition agnostic.

Poggio~\etal~\cite{poggio1987computational} and White~\etal\cite{White2019SharedVA} observe that shapes share abstract similarities while domains vary. Inspired by this observation and unlike most landmark localization methods~\cite{saragih2009face,wei2016convolutional,bulat2017far,wang2019adaptive} our approach models shared \emph{abstract similarities}, i.e., learns together groups of facial landmark semantic groups, Fig~\ref{fig:teaser}, rather than learn landmarks directly. The facial landmark semantic group learning can be shared across domains and definitions. Thus, while image domains and localization definitions vary, learning a single representation for each semantic facial group enables generalization of learning across domains and definitions.

Transformers~\cite{vaswani2017attention} were introduced for natural language processing problems, that model word sequences, e.g., ``[The] [quick] [brown] [fox] ...'', as densely meaningful tokenized vectors. These vectors are initially indexed from a learned embedding matrix which captures each token's definition, learned across training instances, while instance-specific representations are built contextually via a series of attention layers. Inspired by the success of transformers in NLP, the flexible handling of multiple tasks and language domains, Raffel~\etal~\cite{raffel2019exploring}, we consider modeling faces analogously, as a fixed ``sentence'' of tokens representing facial landmark semantic groups. We seek for our model to learn general definitions via semantic group embeddings, as an implicit facial prior, for predicting semantic group information from image feature contexts.

To this end, we design a novel vision transformer (ViT)~\cite{dosovitskiy2020image} architecture for the multiple domain and multi facial landmark definition localization problem. As shown in Figure~\ref{fig:overview}, we first pass the image through our ViT encoder to obtain image feature tokens. These tokens are fed into our novel facial landmark semantic group decoder, which builds contextualized representations of semantic group tokens via cross-attention with image feature tokens and inter-group self-attention. Finally, each facial landmark semantic group vector is passed through definition/dataset specific heads to regress to the final landmarks. Thus, our method treats the facial landmark semantic groups in a general manner, while being capable of predicting landmarks for a variety of domains and definition.

We train our model in a \emph{multi-domain multi-dataset} fashion for small datasets and achieve state-of-the-art performance on \COFW~\cite{burgos2013robust}, a small dataset with only 1345 images and a 29 landmark definition and very competitive performance on \WFLW~\cite{wayne2018lab}. Additionally, we display our method's versatility in adapting to very small (roughly 100 image) datasets of animals (monkeys), caricatures, artwork image domains, and contribute a small novel dataset for pareidolias. We show great improvements via multi-domain multi-dataset learning with ablation, qualitative, and quantitative analysis. To summarize, our contributions include:
\begin{enumerate}
    \item We introduce multi-domain multi-definition learning for the small dataset facial landmark localization problem.
\item We introduce a novel vision transformer encoder-decoder architecture which enables multi-domain (dataset) multi-definition learning via decoding landmark information via shared facial component queries in the decoder.
\item Our method achieves state-of-the-art performance on standard multiple domain facial localization datasets, along with never before seen facial localization domain small datasets, such as, pareidolias.
\end{enumerate}

\section{Related Works}
\label{sec:relatedworks} 
\textit{\textbf{Multi-Domain Learning.}} Multi-domain learning predicts instance labels given both instance features and domain labels, where the goal is to learn a model that improves over a baseline that trains solely on the domain~\cite{Joshi2012MultiDomainLW,Williams2013MultidomainLA}. Similar to our work, several studies utilize multi-domain learning to boost performance on a domain with few labeled examples via concurrent training with a domain with plentiful labels \cite{BenDavid2009ATO,Williams2013MultidomainLA,Dvornik2020SelectingRF,Hoffman2015SimultaneousDT}. Joshi~\etal~\cite{Joshi2012MultiDomainLW} note two approaches for multi-domain learning: domain-specific parameters and modeling inter-domain relations; our approach utilizes both simultaneously. For the image classification task, Dvornik~\etal\cite{Dvornik2020SelectingRF} propose a feature selection approach, while Zheng~\etal~\cite{zheng2021generalfacial} propose a domain confusion loss to encourage the network to learn domain invariant representations for image classification~\cite{Hoffman2015SimultaneousDT}. For facial landmarks, there may exist domain-specific biases in the outputs between domains, so this property is less desired~\cite{Yaniv2019TheFO}. Most similar to our approach, Nam~\etal~\cite{Nam2016LearningMC} introduce multi-domain learning for sequence tracking, where their network shares weights for the bulk of the architecture, with domain specific final layers. Our approach utilizes separate final layers for each domain, while exploiting the relations between domains in our decoder by learning shared representations for facial components.
\\
\textit{\textbf{Multi-Definition Learning.}} The multi-definition problem for facial landmarks solves for inconsistencies between landmark labels to improve model robustness via multi-dataset training \cite{Wu2017LeveragingIA}. Multi-definition learning is similar to multi-domain in the sense that there is a target dataset for which performance is optimized with shared learning from a source dataset \cite{Smith2014CollaborativeFL}. Smith~\etal~\cite{Smith2014CollaborativeFL} propose to predict a super-set of landmark definitions, while Zhu~\etal~\cite{Zhu2014TransferringLA} propose an alignment module to estimate pseudo-labels in schema of a target dataset. Motivated by cross-dataset input variation and definition mismatch, Zhang~\etal~\cite{Zhang2015LeveragingDW} propose an intermediate shape regression module that regresses shared sparse definitions that helps inform final regression to the landmark super-set. Wu~\etal~\cite{Wu2017LeveragingIA} utilize a shared CNN-backbone, prior to dataset/definition specific final direct regression heads. As recent state-of-the-art methods have been heatmap-based, Zhu~\etal~\cite{Zhu2014TransferringLA} propose separate definition-specific heatmap decoders that tightly couple the decoder architectures with output heatmap definitions~\cite{Jin2021SeparableBN}. Our method shares abstract similarity to \cite{Zhang2015LeveragingDW}'s shape regression. We include sparse intermediate predictions that are latent vectors rather than explicit landmarks. Similar to~\cite{Zhang2015LeveragingDW,Wu2017LeveragingIA}, we also employ definition-specific regression heads, see Section~\ref{sec:method}.
\\
\textit{\textbf{CNN and Heatmap-based Landmark Learning.}}
Wei~\etal~\cite{wei2016convolutional} introduce heatmap-based estimation of 2D landmarks for human pose estimation, later Kowalski~\etal~\cite{Kowalski2017DeepAN} adapt it for facial landmarks. While heatmaps provide intrinsic spatial generalization~\cite{Nibali2018NumericalCR}, they induce quantization errors~\cite{jin2021pixel,bulat2021subpixel,lan2021hih}. Stacked hourglass networks~\cite{Newell2016StackedHN,bulat2017far,yang2017stacked} or multi-scale processing ~\cite{sun2019high} are then used for building global context. Jin~\etal~\cite{jin2021pixel} note that connecting CNN features to fully connected layers provides a global predictive capacity that leads to inaccurate predictions due to immediate spatial connections, however, this does lead to more consistent predictions. 

CoordConv~\cite{Liu2018AnIF} connect CNN features with positional information by injecting a fixed spatial bias through two additional image channels that provide global positional information of \{x, y\} coordinates respectively. It was adopted by previous state-of-the-art~\cite{wang2019adaptive} and LAB~\cite{wayne2018lab} to capture global information in CNNs via boundary heatmaps that connect semantic groups of landmarks, e.g., eyes, mouth, etc., into semantically grouped heatmaps on a single global boundary heatmap. Chandran~\etal~\cite{Chandran2020AttentionDrivenCF} propose a hard-attention cropping derived from an initial global pass to consider each semantic region of the face and obtain regional heatmaps for each region for high-resolution images. 
\\
\textit{\textbf{Transformers for Landmark Learning.}} Transformers~\cite{vaswani2017attention} were introduced for vision tasks by DETR~\cite{carion2020end}'s use of a transformer encoder-decoder over CNN-encoded features for the object detection. Vision Transformers (ViT)~\cite{dosovitskiy2020image} show promising performance for vision tasks without the use of CNNs, while DEiT~\cite{touvron2021training} use a CNN for knowledge distillation for further improvements. Swin~\cite{liu2021swin}, inspired by CNN architectures, propose a hierarchically processed shifted window attention approach. However, we adopt a simple vanilla ViT~\cite{dosovitskiy2020image}, and employ a transformer decoder for predicting the latent landmark information for facial semantic group regression.

HiH~\cite{lan2021hih} resolve for heatmap quantization errors and study a CNN-based versus transformer-based heatmap prediction network with a CNN-backbone. LOTR~\cite{watchareeruetai2021lotr} show that transformers can be used to break the direct spatial dependencies induced by CNN-MLP architectures for performant direct regression. They employ a CNN-backbone followed by a transformer-encoder decoder, where the decoder queries correspond to individual landmarks, followed by MLP regression heads. Recently, FarRL~\cite{Bao2021BEiTBP} introduce a BERT/BEiT-like transformer pre-training equivalent on faces, pre-training self-supervisedly on 20 million facial images, and predicting facial landmarks, with heatmap prediction, as one of three facial tasks in a multi-task setup. 
    
Our method combines transformer-based (cross and self-attention) direct regression with the semantic group intuition of LAB, as our novel transformer decoder predicts representations for the semantic groups prior to explicitly regressing landmarks contained in the semantic group. Our method is most similar to LOTR, except that while LOTR is DETR-like~\cite{carion2020end} with its full CNN-backbone, our method is ViT-like, using projection patchification, and no CNN feature backbone. Also, LOTR queries each individual landmark from the encoded image features, whereas our method queries semantic landmark groupings, e.g., nose, for multi-domain/definition learning purposes, and regresses both landmark mean and covariance information.
\\
\textit{\textbf{Multi-Dataset Learning.}} In order to address the small-datasets common among facial landmark problems, several approaches have been devised. These include semi-supervised learning~\cite{honari2018improving,bulat2021subpixel}, self-supervised learning~\cite{zheng2021generalfacial}, and multi-dataset learning~\cite{bulat2021subpixel,zheng2021generalfacial}. Zheng~\etal~\cite{zheng2021generalfacial}, inspired by BERT-inspired~\cite{devlin2018bert} BEiT~\cite{Bao2021BEiTBP}, use self-supervised pre-training techniques to learn general facial representations, employing both a contrastive learning approach using textual labels as well as a masked image prediction methodology on 20 million facial images. Qian\etal~\cite{qian2019aggregation} introduce a synthetic data creation methodology which employs self-supervised learning to translate labeled faces into the style of other images, achieving large performance boosts over purely supervised methods. Jin~\etal~\cite{jin2021separablebatch} introduce cross-protocol network training, where multiple facial landmark datasets are trained simultaneously by sharing a backbone feature encoder and using a different heatmap decoder network for each dataset and thus only shares weights in the CNN feature backbone, but not in the landmark heatmap decoders. Our work is similar to Jin~\etal~in that we train on multi-definition facial landmark datasets as our model's source of additional data. However, rather than decoding each dataset separately, our definition agnostic decoder shares weights across datasets by modeling shared semantic groupings of landmarks.

\section{Method}
\label{sec:method}

Our goal is to create a robust landmark localization solution for small data regime problems, where due to varied circumstances acquiring a large dataset is infeasible or expensive. We approach this problem via a multi-domain multi-definition (MDMD) facial landmark localization formulation with a transformer-based encoder-decoder architecture. The MDMD problem consists of predicting facial landmarks for target image domain(s) or landmark definition(s), while training on one or more source domains/definitions. Our novel transformer-based architecture is shown in Fig~\ref{fig:overview}. We select $n$ image domain datasets, which may have different landmark definitions as our MDMD input. The various landmark definitions map to a standard semantic grouping which we define for each dataset, Sec.~\ref{sec:flsg}. These $n$ domain datasets are then fed into a ViT~\cite{dosovitskiy2020image} encoder that builds image feature representations from the input images, Sec.~\ref{sec:encoder}. Predefined shared semantic group tokens act as a learnt structure ``prior'' to the shared semantic group decoder, that takes as input the encoder's output feature tokens. Then, the decoder builds representations of these definition-agnostic semantic groups by attending to both the image features via cross-attention and the other groups via self-attention, Sec.~\ref{sec:decoder}. Finally, the semantic group tokens -- output of the decoder, each of which individually correspond to a unique set of landmarks (out of $n$), are then fed into regression heads that predict the final landmarks, Sec~\ref{sec:heads}.

\subsection{ViT Encoder}
\label{sec:encoder}
We employ a pre-trained ViT~\cite{dosovitskiy2020image} encoder that is shared across images from all input domains. The model first patchifies the input image to transform the initial image $I \in \mathbb{R}^{224 \times 224 \times 3}$ into a grid $G \in \mathbb{R}^{14 \times 14 \times D}$, with $D=768$, and is then flattened, appended with a global token, and combined with positional encodings to obtain the ViT input tokens $X_{in} \in \mathbb{R}^{(196+1) \times D}$. The input tokens $X$ are then passed through a series of ViT layers consisting of self-attention and MLPs. The final feature tokens are then obtained and passed to the shared decoder to extract landmark information from these generic facial features.

\subsection{Facial Landmark Semantic Group (FLSG)}
\label{sec:flsg}
\begin{wrapfigure}{r}{0.25\textwidth}
  \begin{center}
    \includegraphics[width=0.25\textwidth]{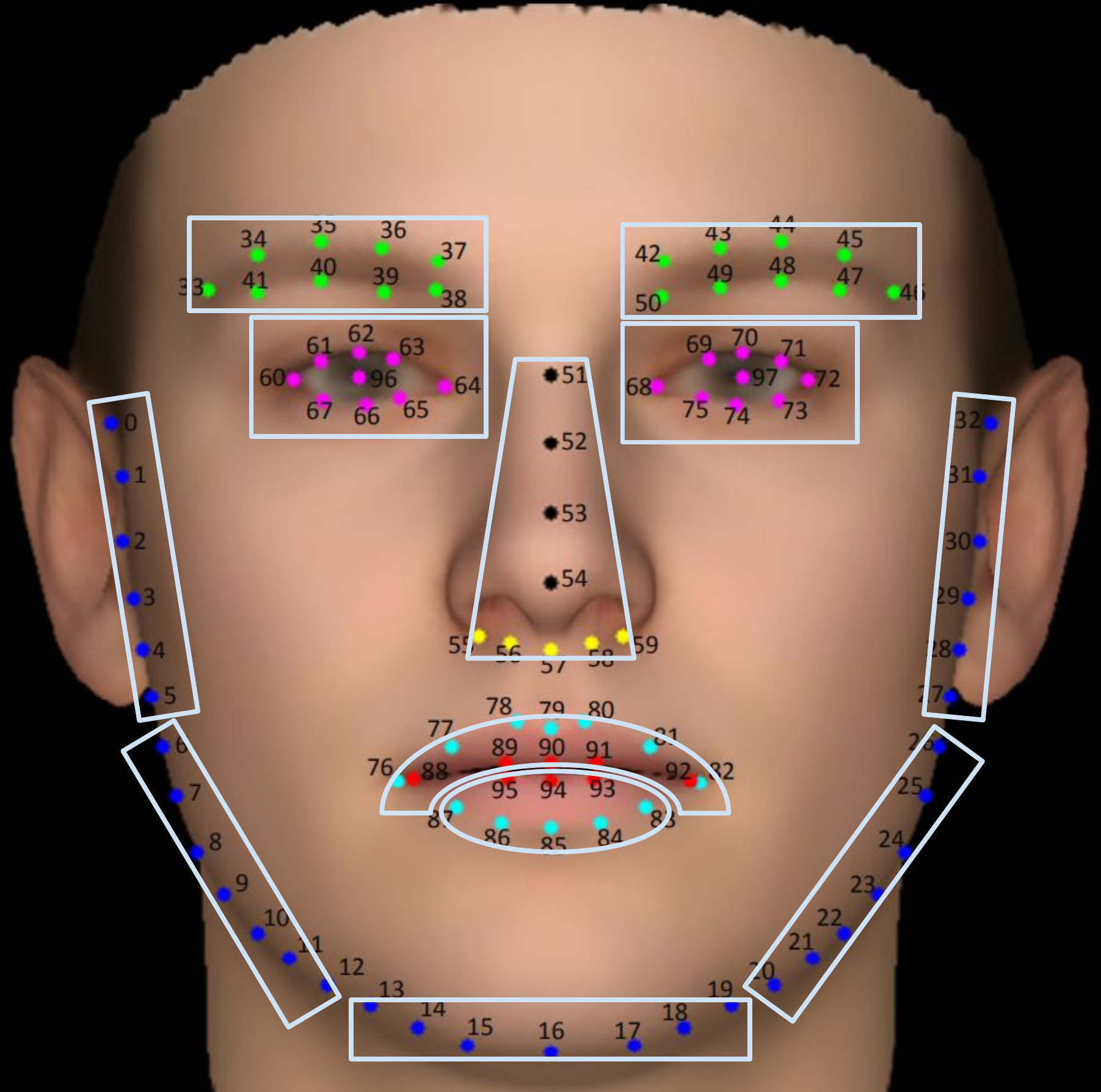}
  \end{center}
  \caption{Facial Landmark Semantic Group. Image source: \cite{wayne2018lab}}
  \label{fig:flsg}
\end{wrapfigure}
In-order to have a universal mechanism for supporting various landmark definitions, we propose a novel facial landmarks semantic group prior. This abstract view of the face leads to definition and domain generalization through the relaxation of strict spatial dependencies, such as in boundary heatmaps~\cite{wayne2018lab}. We divide the facial landmarks for each definition into a set of 12 shared FLSGs, as shown in Figure~\ref{fig:flsg}. FLSGs are modeled as an embedding matrix $F_{in} \in \mathbb{R}^{12\times D}$, where each $F_{in}^{(i)} \in \mathbb{R}^D$ represents learned prior information for a particular semantic group of facial landmarks. The decoder exploits the learned FLSG representations $F_{in}$ that are used to initialize the FLSG tokens that act as input to the decoder. While, different FLSGs may have a different number landmarks depending on the definitions, our model does not explicitly differentiate between them until the final prediction head stage, Section~\ref{sec:heads}. Thus, our FLSGs unlock MDMD learning via a standard general facial representation.

\subsection{Definition Agnostic Decoder}
\label{sec:decoder}

We want FLSGs to collect information from the image features that it deems relevant (cross-attention) and also collect information about its context wrt other FLSGs (self-attention). We achieve this via a novel definition agnostic decoder, where given initial FLSG tokens $F_{in} \in \mathbb{R}^{12\times D}$ and the encoded image feature tokens $X_{out} \in \mathbb{R}^{(196+1)\times D}$, the decoder seeks to infuse the ``structured prior'' FLSG tokens with information from both the input image and other FLSG tokens. The decoder is composed of three decoder blocks that consist of self- and cross-attention. Each decoder block contains cross-attention in which the FLSG tokens act as ``queries'' and the image features as ``keys'' and ``values''~\cite{vaswani2017attention}. This is followed by self-attention, where the FLSG tokens can perform message passing. Explicitly, given the input FLSG tokens, $F_{in}$, each decoder block is as follows: 
%
\begin{align}
    F_{hidden}^1 &= \mathsf{MHCA}(\mathsf{LN}(F_{in}), \mathsf{LN}(X_{out}))) + F_{in}\\
    F_{hidden}^2 &= \mathsf{MHSA}(\mathsf{LN}(F_{hidden}^1)) + F_{hidden}^1\\ 
    F_{out} &= \mathsf{FFN}(\mathsf{LN}(F_{hidden}^2))
\end{align}
where $\mathsf{MHCA}$, $\mathsf{MHSA}$, $\mathsf{LN}$, and $\mathsf{FFN}$ are the standard transformer multi-head cross-attention, multi-head self-attention, layer normalization, and feed-forward network respectively~\cite{vaswani2017attention}. Thus, decoder layers infuse the FLSG tokens with image feature information as well as inter-FLSG contextual information, so that they contain information pertaining to localizing the landmarks contained in the given semantic group. The final FLSG tokens $F_{out} \in \mathbb{R}^{12\times D}$, output by decoder, are then plugged into the definition specific prediction heads.

\subsection{Definition/Domain-Specific Prediction Heads}
\label{sec:heads}
Finally, we employ definition/domain specific prediction heads, that directly regress the landmarks that correspond to each FLSG. An image $I_j$ is provided to our model, where $j$ is the dataset (definition) index. The dataset index is simply used to route the FLSG tokens to the head that corresponds to that dataset (see pseudocode in the supplementary). Each dataset's landmark head, regresses from an FLSG vector $F_{out}^i \in \mathbb{R}^D$ via two-layer ${\mathsf{MLP_{lm}}}^i_j$ to output landmarks $L^i_j \in \mathbb{R}^{N_j^i\times2}$, where $N_j^i$ is the number of landmarks for the $i$th FLSG and the $j$th dataset.

Rather than predict landmarks alone, following Kumar~\etal~\cite{kumar2020luvli}, we also predict the covariance information via a Cholesky estimation head, ${\mathsf{MLP_{chol}}}^i_j$, obtaining a second output corresponding to each FLSG, $C^i_j \in \mathbb{R}^{N_j^i\times3}$, corresponding to the parameters of the Cholesky factorization of a predicted covariance matrix. While Kumar~\etal's Cholesky estimation network regresses from the latent bottleneck vector of their CNN, and use heatmap-based prediction for the mean estimate, supervising outputs from several stacked hourglass layers of their DU-NET~\cite{Tang2020TowardsEU}, we utilize a shared FLSG vector to predict both mean and covariance information. The final minimization loss function we use to train our model is as follows:
\begin{align}
\mathcal{L}_{\mathsf{MDMD}} = \frac{1}{|\mathsf{FLSG}|}\sum_{i=1}^{|\mathsf{FLSG}|}\bigg{[} \frac{1}{N^i_j}\sum_{k=1}^{N^i_j} \mathcal{L}_{lll} (L^i_j, C^i_j, {L_{GT}}_j^i)_k\bigg{]}
\end{align}
Here, ${L}_{lll}$ is the Laplacian log-likelihood (see supplementary) and ${L_{GT}}_j^i$ is the ground truth landmarks.

\section{Experiments and Results}
\label{sec:results}

\begin{figure}[t]
\center
  \includegraphics[width=\textwidth]{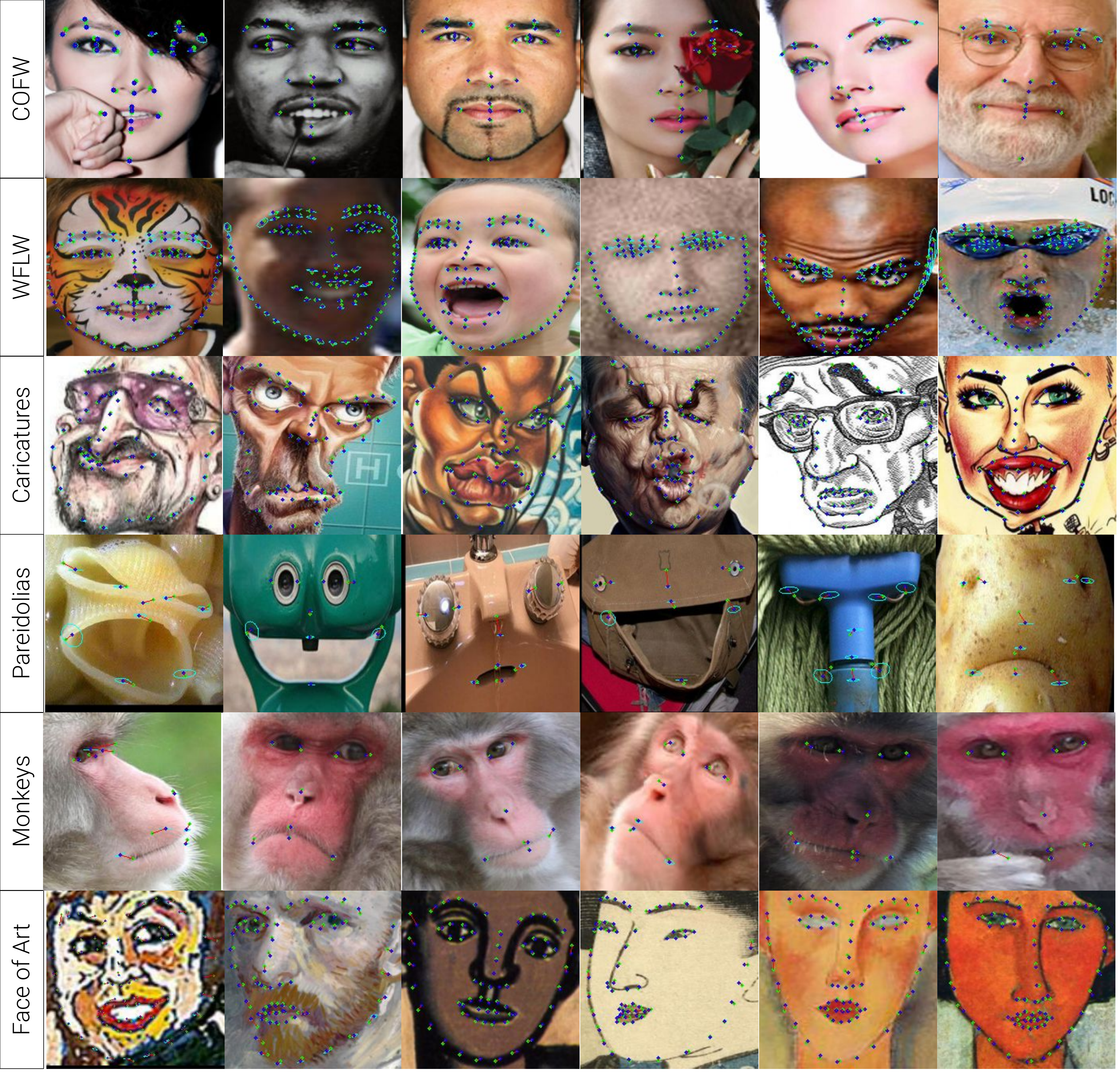}
  \caption{Qualitative results for our method across several datasets. Key: {\color{green}GT landmarks}, {\color{blue}predicted landmarks}, {\color{red}error vectors}, {\color{teal}uncertainty estimation}}
  \label{fig:results}
\end{figure}

We evaluate our model's multi-domain and multi-definition learning capabilities on novel domains with small datasets: \AnimalWeb ~\cite{Khan2020AnimalWebAL}, \ArtisticFaces ~\cite{Yaniv2019TheFO}, \CaricatureFace ~\cite{Zhang2021LandmarkDA} and \PARE~dataset [New], as well as standard benchmark datasets: \COFW~\cite{burgos2013robust} and \WFLW~\cite{wayne2018lab} (and \TW~\cite{sagonas2013300}, \LAPA~\cite{liu2019high}), see supplementary material for details on datasets.

For each experiment, we report normalized mean error (NME) with inter-ocular normalization as well as inter-pupil, where comparison necessitates. Additionally, we report Area Under The Curve (AUC) and FR (Failure Rate) scores, considering a failure as mean NME greater than 10\% for a given face. While $256\times256$ input crops are most commonly used ~\cite{qian2019aggregation}, our ViT~\cite{dosovitskiy2020image} encoder was pre-trained with $224\times224$ input crops, that we adopt. All models are trained with the Adam\cite{kingma2014adam} optimizer with learning rate $1e^{-4}$ and linear learning rate decay. For each experiment, we consider performance for our model training with a single domain and definition, and then compare its performance when training with an additional dataset in the multi-domain and multi-definition fashion. In order to train concurrently across datasets, for each mini-batch, we uniformly sample a dataset from which we draw batch samples. We include additional implementation details, including augmentation strategy, in the supplementary materials. In the following, we discuss various qualitative (Figure~\ref{fig:results}), and quantitative results on various datasets: 

\subsection{\COFW~\cite{burgos2013robust}}
We evaluate our method on the \COFW~dataset that contains 1,345 training images, and 500 testing images. We note that among standard benchmark datasets, \COFW~is most similar to our problem for its unique 29 landmark definition as well as its relatively small size. We train our model with two settings: \COFW, and \COFW~concurrently trained with \LAPA. We evaluate our model with inter-pupil normalization, following ~\cite{wang2019adaptive,huang2021leveraging}, surpassing state-of-the-art, Table~\ref{tbl:cofw_comparison}. We also note that for each dataset on which we train our model, concurrent training with a larger dataset shows significant performance improvements.

\begin{table}[t]
\center
\caption{Comparison against SOTA for \COFW~\cite{burgos2013robust}}
\begin{tabular}{|l|l|l|l|}
\hline
Method    & \multicolumn{1}{c|}{NME$_{ip}$(\%)}  & \multicolumn{1}{c|}{FR$_{10\%}$}    & \multicolumn{1}{c|}{AUC$_{10\%}$} \\ \hline
Wing~\cite{feng2018wing}      & \multicolumn{1}{c|}{5.44} & \multicolumn{1}{c|}{3.75} & \multicolumn{1}{c|}{-} \\  
DCFE~\cite{Valle2018ADC}     & \multicolumn{1}{c|}{5.27} & \multicolumn{1}{c|}{7.29} & \multicolumn{1}{c|}{35.86} \\ 
AWing~\cite{wang2019adaptive}     & \multicolumn{1}{c|}{4.94} & \multicolumn{1}{c|}{.99} & \multicolumn{1}{c|}{48.82} \\ 
ADNet~\cite{huang2021leveraging}      & \multicolumn{1}{c|}{4.68} & \multicolumn{1}{c|}{.59} & \multicolumn{1}{c|}{53.17} \\ \hline 
MDMD Base & \multicolumn{1}{c|}{4.82}&\multicolumn{1}{c|}{\textbf{.39}}&\multicolumn{1}{c|}{51.84} \\ 
MDMD w/LaPa & \multicolumn{1}{c|}{\textbf{4.65}}&\multicolumn{1}{c|}{.59} &\multicolumn{1}{c|}{\textbf{53.49}} \\ \hline 
\end{tabular}
\label{tbl:cofw_comparison}
\end{table}

\subsection{\WFLW~\cite{wayne2018lab}}
We further evaluate our method on the \WFLW~dataset, which consists of 7,500 training images and 2,500 testing images, with a 98 landmark definition. We train our model in the multi-definition manner with two settings: \WFLW~and \WFLW~ concurrently with \LAPA, where \LAPA~presents 19,000 faces with a 106 landmark definition. As \TW~and \COFW~are relatively small with 3837 and 1345 training faces respectively, we do not consider the concurrently training with these smaller datasets, as this runs contrary to our goal of boosting performance from training with larger datasets. We compare our results with other methods for NME, FR, and AUC on the full test set along with subsets which test for robustness on large poses, expression, illumination, make-up, occlusion, and blur, in Table~\ref{tbl:wflw_comparison}. Our method outperforms all previous state-of-the-art methods for overall scores aside from two concurrent works ~\cite{bulat2021subpixel,zheng2021generalfacial}. Our method also achieves SOTA performance compared to previously reported methods for the majority of subsets for NME, FR, and AUC. See qualitative comparisons for our method in Figure~\ref{fig:results}.

\begin{table}[] 
\center
\caption{Comparison against SOTA for \WFLW~\cite{wayne2018lab}. *Concurrent works, Key: {\color{green}best}, {\color{blue}second}}
\begin{adjustbox}{width=\linewidth}
\centering
\begin{tabular}{|c|l|c|c|c|c|c|c|c|}
\hline
Metric               & \multicolumn{1}{c|}{Method} & \multicolumn{1}{c|}{Testset} & \multicolumn{1}{c|}{\begin{tabular}[c]{@{}c@{}}Pose\\ Subset\end{tabular}} & \multicolumn{1}{c|}{\begin{tabular}[c]{@{}c@{}}Expression\\ Subset\end{tabular}} & \multicolumn{1}{c|}{\begin{tabular}[c]{@{}c@{}}Illumination\\ Subset\end{tabular}} & \multicolumn{1}{c|}{\begin{tabular}[c]{@{}c@{}}Make-up\\ Subset\end{tabular}} & \multicolumn{1}{c|}{\begin{tabular}[c]{@{}c@{}}Occlusion\\ Subset\end{tabular}} & \multicolumn{1}{c|}{\begin{tabular}[c]{@{}c@{}}Blur \\ Subset\end{tabular}} \\ \hline

\multirow{13}{*}{NME(\%)}
& ESR~\cite{Cao2012FaceAB} & 11.13 & 25.88 & 11.47 & 10.49 & 11.05 & 13.75 & 12.20 \\ 
& SDM~\cite{Xiong2013SupervisedDM} & 10.29 & 24.10 & 11.45 & 9.32 & 9.38 & 13.03 & 11.28 \\ 
& CFSS~\cite{Zhu2015FaceAB} & 9.07 & 21.36 & 10.09 & 8.30 & 8.74 & 11.76 & 9.96 \\ 
& DVLN~\cite{Wu2017LeveragingIA} & 6.08 & 11.54 & 6.78 & 5.73 & 5.98 & 7.33 & 6.88 \\ 
& LAB~\cite{wayne2018lab} & 5.27 & 10.24 & 5.51 & 5.23 & 5.15 & 6.79 & 6.12 \\ 
& Wing~\cite{feng2018wing} & 5.11 & 8.75 & 5.36 & 4.93 & 5.41 & 6.37 & 5.81 \\ 
& DeCaFA~\cite{Dapogny2019DeCaFADC} & 4.62 & 8.11 & 4.65 & 4.41 & 4.63 & 5.74 & 5.38 \\ 
& AWing~\cite{wang2019adaptive} & 4.36 & 7.38 & 4.58 & 4.32 & 4.27 & 5.19 & 4.96 \\ 
& LUVLi~\cite{kumar2020luvli} & 4.37 & - & - & - & - & - & - \\ 
& AWing~\cite{wang2019adaptive}      & 4.36 & 7.38 & 4.58 & 4.32 & 4.27 & 5.19 & 4.96 \\ 
& HiH~\cite{lan2021hih} & 4.18  & 7.20  & {\color{blue}4.19}  & 4.45  & 3.97  & 5.00  & 4.81 \\ 
& ADNet~\cite{huang2021leveraging}       & 4.14 & 6.96 & 4.38 & 4.09 & 4.05 & 5.06 & 4.79 \\ 
& FaRL~\cite{zheng2021generalfacial}*       & {\color{blue}3.96} & {\color{blue}6.91} & 4.21 & 3.97 & {\color{green}3.80} & {\color{green}4.71} & {\color{blue}4.57} \\ 
& SH-FAN Base~\cite{bulat2021subpixel}*& 4.20 & -    & -    & -    & -    & -    &      \\ 
& SH-FAN~\cite{bulat2021subpixel}*     & {\color{green}3.72} & -    & -    & -    & -    & -    &      \\ 
& \MDMDBase*      & 4.06 & 7.11 & 4.21 & {\color{blue}3.88} & 4.04 & 4.86 & 4.63 \\ 
& \MDMD*  & 3.97  & {\color{green}6.90}  & {\color{green}4.11}  & {\color{green}3.80}  & {\color{blue}3.90}  & {\color{blue}4.78}  & {\color{green}4.49} \\ \hline 

\multirow{13}{*}{FR$_{10}$(\%)}
& ESR~\cite{Cao2012FaceAB} & 35.24 & 90.18 & 42.04 & 30.80 & 38.84 & 47.28 & 41.40 \\ 
& SDM~\cite{Xiong2013SupervisedDM}        & 29.40 & 84.36 & 33.44 & 26.22 & 27.67 & 41.85 & 35.32 \\ 
& CFSS~\cite{Zhu2015FaceAB}       & 20.56 & 66.26 & 23.25 & 17.34 & 21.84 & 32.88 & 23.67 \\ 
& DVLN~\cite{Wu2017LeveragingIA}       & 10.84 & 46.93 & 11.15 & 7.31 & 11.65 & 16.30 & 13.71 \\ 
& LAB~\cite{wayne2018lab}        & 7.56 & 28.83 & 6.37 & 6.73 & 7.77 & 13.72 & 10.74 \\ 
& Wing~\cite{feng2018wing}       & 6.00 & 22.70 & 4.78 & 4.30 & 7.77 & 12.50 & 7.76 \\ 
& DeCaFA~\cite{Dapogny2019DeCaFADC}     & 4.84 & 21.40 & 3.73 & 3.22 & 6.15 & 9.26 & 6.61 \\ 
& AWing~\cite{wang2019adaptive}      & 2.84 & 13.50 & 2.23 & 2.58 & 2.91 & 5.98 & 3.75 \\ 
& LUVLi~\cite{kumar2020luvli}      & 3.12 & - & - & - & - & - & - \\ 
& HiH~\cite{lan2021hih} & 2.84 & 14.41 & 2.55 & 2.15 & {\color{green}1.46} & 5.71 & 3.49 \\
& ADNet~\cite{huang2021leveraging}       & 2.72  & {\color{blue}12.72} & 2.15 & 2.44 & {\color{blue}1.94} & 5.79 & 3.54 \\ 
& FaRL~\cite{zheng2021generalfacial}*       & {\color{blue}1.76} & - & - & - & - & - & - \\ 
& SH-FAN~\cite{bulat2021subpixel}*     & {\color{green}1.55} & -    & -    & -    & -    & -    & -     \\ 
& \MDMDBase*       & 2.63 & 14.11 & {\color{blue}1.91} & {\color{blue}1.71} & 2.43 & {\color{blue}4.89} & {\color{blue}2.98}     \\
& \MDMD*  & 2.2  & {\color{green}11.96} & {\color{green}1.27}  & {\color{green}1.58}  & {\color{green}1.46}  & {\color{green}4.35}  & {\color{green}2.59}  \\ \hline

\multirow{3}{*}{AUC$_{10\%}$}
& ESR~\cite{Cao2012FaceAB} & 0.2774 & 0.0177 & 0.1981 & 0.2953 & 0.2485 & 0.1946 & 0.2204 \\ 
& SDM~\cite{Xiong2013SupervisedDM} & 0.3002 & 0.0226 & 0.2293 & 0.3237 & 0.3125 & 0.2060 & 0.2398 \\ 
& CFSS~\cite{Zhu2015FaceAB} & 0.3659 & 0.0632 & 0.3157 & 0.3854 & 0.3691 & 0.2688 & 0.3037 \\ 
& DVLN~\cite{Wu2017LeveragingIA} & 0.4551 & 0.1474 & 0.3889 & 0.4743 & 0.4494 & 0.3794 & 0.3973 \\ 
& LAB~\cite{wayne2018lab} & 0.5323 & 0.2345 & 0.4951 & 0.5433 & 0.5394 & 0.4490 & 0.4630 \\ 
& Wing~\cite{feng2018wing} & 0.5504 & 0.3100 & 0.4959 & 0.5408 & 0.5582 & 0.4885 & 0.4918 \\ 
& DeCaFA~\cite{Dapogny2019DeCaFADC} & 0.5630 & 0.2920 & 0.5460 & 0.5790 & 0.5750 & 0.4850 & 0.4940 \\ 
& AWing~\cite{wang2019adaptive} & 0.5719 & 0.3120 & 0.5149 & 0.5777 & 0.5715 & 0.5022 & 0.5120 \\ 
& LUVLi~\cite{kumar2020luvli} & 0.5770 & - & - & - & - & - & - \\ 
& HiH~\cite{lan2021hih} & 0.597 & 0.342 & {\color{blue}0.590} & 0.606 & {\color{blue}0.604} & 0.527 & 0.549 \\
& ADNet~\cite{huang2021leveraging} & 0.6022 & {\color{green}0.3441} & 0.5234 & 0.5805 & 0.6007 & 0{\color{blue}.5295} & {\color{blue}0.5480} \\ 
& FaRL~\cite{zheng2021generalfacial}* & {\color{blue}0.6116} & - & - & - & - & - & - \\ 
& SH-FAN~\cite{bulat2021subpixel}* & {\color{green}.6310} & -    & -    & -  & - & -      & -     \\ 
& \MDMDBase* & .6010 & .3316 & .5870 & {\color{blue}.6179} & .5998 & .4978  & .5476      \\
& \MDMD*  & .6083 & {\color{blue}.3438} & {\color{green}.5933} & {\color{green}.6252} & {\color{green}.6127} & {\color{green}.5354} & {\color{green}.5582} \\ \hline

\end{tabular}
\end{adjustbox}
\label{tbl:wflw_comparison}
\end{table}

\subsection{Small Dataset Experiments}
We consider our methods performance for small datasets of novel domains and landmark definitions. For each of these experiments, we train both a baseline model on the small dataset only as well as a multi-domain and multi-definition model, for which we employ the moderately sized \TW~dataset. Per Williams~\etal~\cite{Williams2013MultidomainLA}, the goal of multi-domain learning is to show improvement over a single domain baseline. While for previous experiments, our focus was primarily on how our method compares with previous methods, here, we compare against a baseline single-domain training. Where applicable we draw rough comparisons with previous works for these datasets. We report relative NME, FR, and AUC for all small dataset experiments in Table~\ref{tbl:small_dataset_comparison}. We observe large performance gains for each dataset through the generalized learning via our multi-domain and multi-definition approach.

\begin{table}
\caption{Evaluation of multi-domain and multi-definition learning capabilities across small datasets for novel domains and landmark definitions} 
\center
\begin{tabular}{|l|l|c|c|c|c|}
\hline
Dataset                            & Method                                      & NME$_{ic}$ & FR$_{10\%}$ & AUC$_{10\%}$ & \# Landmarks \\ \hline
\multirow{2}{*}{\AnimalWeb ~\cite{Khan2020AnimalWebAL}}       &\multicolumn{1}{|c|}{MDMD Base}   & \multicolumn{1}{c|}{6.88} & \multicolumn{1}{c|}{\textbf{15.15}} & \multicolumn{1}{c|}{.4233} & 9 \\
                                             &\multicolumn{1}{|c|}{MDMD w/300W} & \multicolumn{1}{c|}{\textbf{6.55}} & \multicolumn{1}{c|}{\textbf{15.15}} & \multicolumn{1}{c|}{\textbf{.4388}} & 9 \\ \hline
\multirow{2}{*}{\ArtisticFaces  ~\cite{Yaniv2019TheFO}}
                                            &\multicolumn{1}{|c|}{MDMD Base}   &  4.46 &  2.08 & .5549 & 68\\
                                            &\multicolumn{1}{|c|}{MDMD w/300W} & \multicolumn{1}{c|}{\textbf{3.75}} & \multicolumn{1}{c|}{\textbf{0.0}} & \multicolumn{1}{c|}{\textbf{.63}} & 68\\ \hline
\multirow{2}{*}{\CaricatureFace  ~\cite{Zhang2021LandmarkDA}} &\multicolumn{1}{|c|}{MDMD Base}   & 7.81 &  19.04 & .2941   & 68\\
                                             &\multicolumn{1}{|c|}{MDMD w/300W} & \multicolumn{1}{c|}{\textbf{5.85}} & \multicolumn{1}{c|}{\textbf{6.41}} & \multicolumn{1}{c|}{\textbf{.4357}} & 68\\ \hline
\multirow{2}{*}{\PARE}                                      &\multicolumn{1}{|c|}{MDMD Base}   &9.12&28.0&.2365& 9 \\
                                            &\multicolumn{1}{|c|}{MDMD w/300W} & \multicolumn{1}{c|}{\textbf{8.59}} & \multicolumn{1}{c|}{\textbf{22.0}} & \multicolumn{1}{c|}{\textbf{.2871}} & 9 \\ \hline
\end{tabular}
\label{tbl:small_dataset_comparison}
\end{table}
%
\AnimalWeb ~\cite{Khan2020AnimalWebAL}\
\\
While the \AnimalWeb ~\cite{Khan2020AnimalWebAL} dataset features 21,900 animal faces across 334 species, we select a single specie, the Japanese Macaque, containing 133 examples which we split into 100 training and 33 testing monkey faces for our experiment. We train jointly between \TW~and the monkey domains and definitions. We note that the animals are labeled with 9 landmarks, while \TW~is labeled with 68. For comparison against previous work, we cannot compare directly, as Khan~\etal~\cite{Khan2020AnimalWebAL} train on a variety of species on a dataset with significantly large amount of training data. Furthermore, their scores represent a wide variety of animals, while ours are a subset of just one animal. Nevertheless, we report our scores for ballpark comparison in Table~\ref{tbl:animal_comparison} (left).

\begin{table}[t]
\caption{Comparison against previous work for \AnimalWeb~\cite{Khan2020AnimalWebAL} (left) and \ArtisticFaces~\cite{Yaniv2019TheFO} (middle). Following Khan, NME scores for \AnimalWeb~are normalized by bounding box size. Comparison against previous work for \CaricatureFace~\cite{Zhang2021LandmarkDA}(right).}
\center
\begin{tabular}{|l|c|}
\hline
\multicolumn{1}{|l|}{Method} & \multicolumn{1}{l|}{NME$_{box}$} \\ \hline
Khan \etal ~\cite{Khan2020AnimalWebAL} & 5.23        \\ \hline
MDMD Base                    & 3.66                 \\ \hline
MDMD 300W                  & \textbf{3.44}                 \\ \hline
\end{tabular}
\begin{tabular}{|l|c|}
\hline
\multicolumn{1}{|l|}{Method} & \multicolumn{1}{l|}{NME$_{ic}$} \\ \hline
Yaniv \etal ~\cite{Yaniv2019TheFO}   & 4.52{\color{red}$^2$}   \\ \hline
MDMD Base                    & 4.46           \\ \hline
MDMD 300W                  & \textbf{3.72}          \\ \hline
\end{tabular}
\begin{tabular}{|l|c|c|}
\hline
\multicolumn{1}{|l|}{Method} & \multicolumn{1}{l|}{NME} & Trn Imgs \\ \hline
Zhang\etal ~\cite{Zhang2021LandmarkDA} & \textbf{5.83}       & 6,420                    \\ \hline
MDMD Base                    & 7.81  & 148                     \\ \hline
MDMD 300W                  & 5.85 & 148                     \\ \hline
\end{tabular}
\label{tbl:animal_comparison}
\end{table}
\ArtisticFaces~\cite{Yaniv2019TheFO}\
\\
We compare our model's performance on `faces in artworks' domain, utilizing the \ArtisticFaces~\cite{Yaniv2019TheFO} dataset, consisting of 160 faces, 10 faces per 16 artists, with a \TW-like 68 landmark definition. While Yaniv~\etal~\cite{Yaniv2019TheFO} utilize an elaborate geometric-aware and style transfer to augment \TW~images for training, and perform generalization to the artworks domain for testing, we split each artist by taking the first 7 image indices for training with the other 3 for testing. \footnote{We compare our results against previous work, with a caveat that our evaluation is on a subset of the dataset rather than the full dataset, and achieve SOTA performance for the \ArtisticFaces~dataset, as shown in Table~\ref{tbl:animal_comparison} (right).}
\\
\CaricatureFace~\cite{Zhang2021LandmarkDA}\
\\
Zhang~\etal~\cite{Zhang2021LandmarkDA} introduce an interesting problem of localizing landmarks on the domain of human caricatures, introducing the \CaricatureFace~dataset that they train on 6,240 images and test on 1,560. Rather than train on the full set, we simulate the small dataset problem for this novel domain by taking the first 148 images for training and evaluating on the full test set. Similar to \ArtisticFaces~\cite{Yaniv2019TheFO}, \CaricatureFace~\cite{Zhang2021LandmarkDA} uses the same \TW~landmark definition. We compare our NME scores trained on 40X less data from the caricature domain, and achieve slightly lower performance than Zhang~\etal, as shown in Table~\ref{tbl:animal_comparison}.
\\
\PARE
\\
Finally, we consider a unique dataset of illusory faces, also known as pareidolias, that we obtained from~\cite{wardle2022illusory}, and labeled 150 images with 9 landmarks each. This domain is particularly interesting, as the face pictures are only abstractly similar to the human faces from the \TW~dataset with which the model trains concurrently. As shown in Table~\ref{tbl:small_dataset_comparison}, performance greatly improves with multi-domain and multi-definition learning.

\begin{figure}[t]
\centering
  \includegraphics[width=\textwidth]{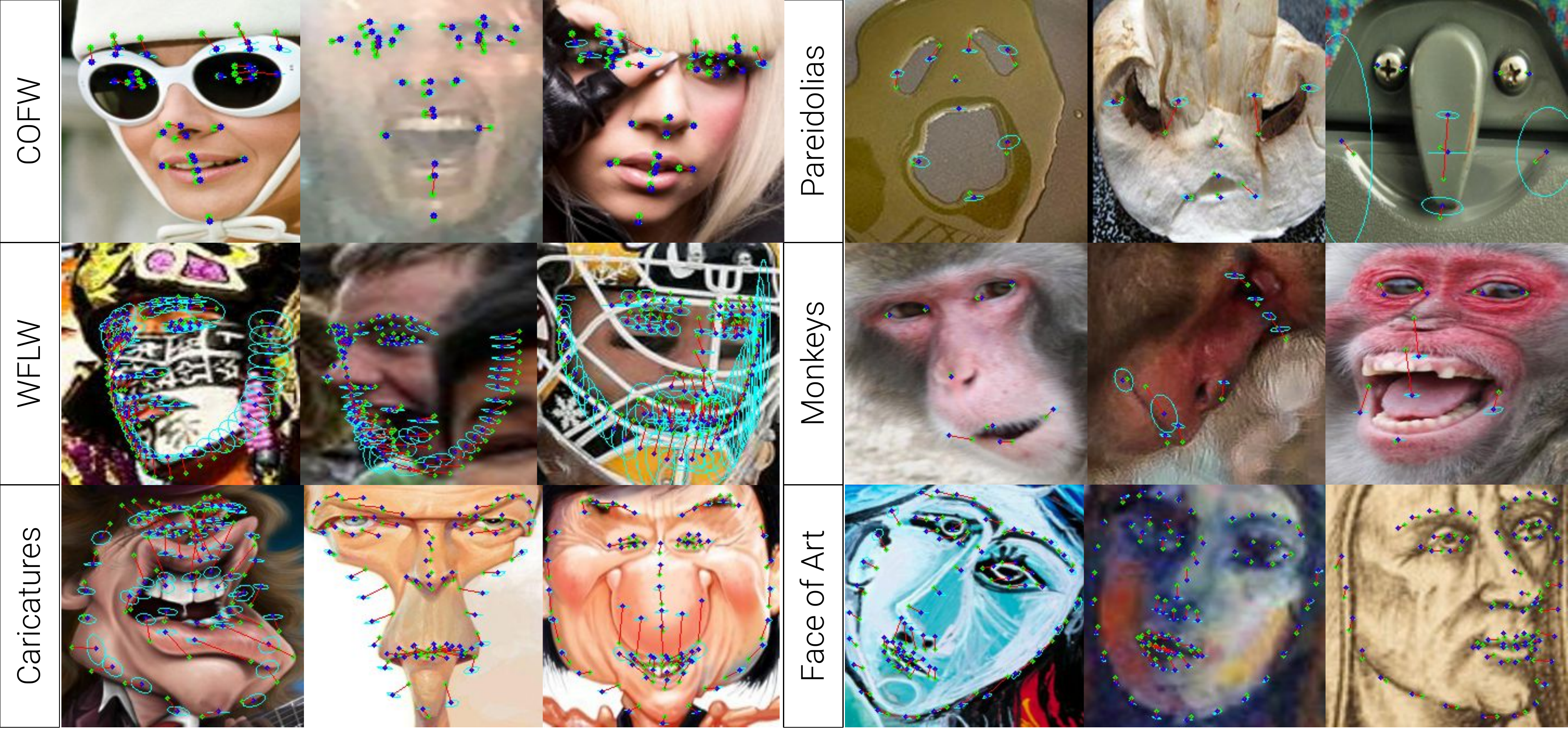}
  \caption{Qualitative results for displaying the limitations our method across several datasets. Key: {\color{green}GT landmarks}, {\color{blue}predicted landmarks}, {\color{red}error vectors}, {\color{teal}uncertainty estimation}}
  \label{fig:limitations}
\end{figure}

\begin{table}[t]
\caption{Ablation studies.}
\center
\begin{tabular}{|l|c|c|c|l|}
\hline
\multicolumn{1}{|l|}{Method} & \multicolumn{1}{l|}{NME$_{ip}$} & \multicolumn{1}{l|}{FR$_{10\%}$} & \multicolumn{1}{l|}{AUC$_{10\%}$} \\ \hline
MDMD Single w/Euclidean loss   & 4.90 & .79 &  .5100 \\ \hline 
MDMD Single w/landmark tokens  & 4.73 & \textbf{.39} & .5278 \\ \hline 
MDMD Single                    & 4.82 & \textbf{.39} &   .5184 \\ \hline
MDMD w/LaPa                     & \textbf{4.64} &  .59 & \textbf{.5349} \\ \hline
\end{tabular}
\label{tbl:ablation}
\end{table}

\subsection{Ablation Analysis}
\label{sec:analysis}

In addition to training with and without an additional dataset, we perform ablation studies for a several architectural components of our model. For each study, we test performance on the \COFW~dataset alone. First, we remove our facial landmark semantic grouping tokens from our decoder, and replace them with individual landmark tokens. Next, we train with simple Euclidean loss, rather than Lapalacian log-likelihood. We show our comparisons for against the baseline model in Table~\ref{tbl:ablation}.
\\
\textit{Small Datasets with v. without \TW~\cite{sagonas2013300}}. We compare our model's performance with and without an additional dataset when training on small datasets of novel domains and definitions. We observe that for each dataset, training without the additional data leads to severe performance reductions, Table~\ref{tbl:small_dataset_comparison}. Thus, we conclude that our multi-domain and multi-domain learning strategy is effective at exploiting additional labeled data for small datasets of novel domains and definitions.
\\
\textit{Laplacian Log-Likelihood v. Euclidean Loss.}
To evaluate the effectiveness of our Laplacian log-likelihood training objective, we compare against a simple baseline of Euclidean distance loss. We train our model on COFW~\cite{burgos2013robust} and show that performance severely deteriorates when we use Euclidean loss, Table~\ref{tbl:ablation}.
\\
\textit{Facial Landmark Semantic Group (FLSG) v. Explicit Landmark Modeling.}
Lastly, we seek to evaluate the effectiveness of our FLSG modeling when compared to a simple baseline of modeling each landmark with its own token. As our MDMD method relies on FLSGs to accomplish multi-definition learning, and thus, cannot be removed while still accomplishing the same task, we instead consider its effectiveness when training with a single dataset, COFW~\cite{burgos2013robust}. We observe a decrease in performance when training with the FLSG in the standard scenario of a single dataset, Table~\ref{tbl:ablation}. However, this decrease is overcome by multi-dataset learning. Thus, FLSG acts as a strategy for achieving performance gains in the multi-domain/definition scenario, while landmark queries was better for the single dataset case, in this case.

\section{Limitations and Conclusion}
\label{sec:conclusion}
We introduced a method for multi-domain and multi-definition landmark localization, that employs a transformer that models facial landmark semantic groups (FLSGs) as opposed to individual landmarks, in-order to share learning across domains and definitions. Our method achieves state-of-the-art performance, as well as successfully improves over baselines of single-domain learning for both small and large datasets. We note however, that our model still struggles with certain difficult circumstances, such as extreme pose and occlusions, as shown in Figure~\ref{fig:limitations}. Another limitation is extremely deformed face shapes, for example, the middle caricature face, Fig.~\ref{fig:limitations}.

We also note that FLSGs aid in multi-definition learning, but hurt performance for the single dataset scenario (Table~\ref{tbl:ablation}). Thus, in the future, we want to explore exploitation of explicit landmark modeling jointly with FLSGs to obtain the best of both worlds. We also note that the proposed FLS-Grouping (after several test permutations) works well for all domains/definitions and helps with generalization, in the future we want to explore domain/definition specific grouping. Additional future work may consider extending these ideas to multi-task learning, temporal modeling, or toward zero-shot and few-shot learning.

\bibliographystyle{splncs04}
\bibliography{main}

\clearpage

\appendix

\section{Dataset Details}
\paragraph{Standard Benchmark Datasets.} See Qian~\etal~\cite{qian2019aggregation} for a detailed description of the \WFLW, \COFW, and \TW~datasets, and Liu \etal~\cite{liu2019high} for \LAPA.
\begin{enumerate}
    \item \WFLW~\cite{wayne2018lab}: 7,500 training faces, 98 landmarks
    \item \LAPA~\cite{liu2019high}: 18,176 training faces, 106 landmarks
    \item \COFW~\cite{burgos2013robust}: 1345 training faces, 29 landmarks
    \item \TW~\cite{sagonas2013300}: 3837 training faces, 68 landmarks
\end{enumerate}
\paragraph{Non-Standard Datasets.} While the \AnimalWeb~\cite{Khan2020AnimalWebAL} and \CaricatureFace~\cite{Zhang2021LandmarkDA} datasets contain larger numbers of images, in this study, for the purpose of evaluating our method's performance for novel domains with small datasets, we only consider a single animal from \AnimalWeb, the Japanese macaque, for its greater visual similarity with human faces, as well as the first 148 images of \CaricatureFace. Additionally, we utilize a small unlabeled dataset of 150 \emph{in-the-wild} illusory faces~\cite{wardle2022illusory}, called pareidolias. We label the bounding boxes in addition to a 9 landmark definition, following \AnimalWeb ~\cite{Khan2020AnimalWebAL}, and refer to this dataset of PAREeidolias as the \PARE~dataset. We will release the include the GT landmarks and images indices from the dataset used for PARE.
\begin{enumerate}
    \item \AnimalWeb ~\cite{Khan2020AnimalWebAL}: 17,520 (80\% of 21,900) training faces, 9 landmarks, 334 animal species
    \item \ArtisticFaces ~\cite{Yaniv2019TheFO}: 160 faces, 68 landmarks, 16 artists, 10 per artist
     \item \CaricatureFace ~\cite{Zhang2021LandmarkDA}: 6,240 (80\% of 7,800) training faces, 68 landmarks
    \item \PARE~dataset [New]: 150 ``faces'', 9 landmarks
\end{enumerate}

\section{Laplacian Log-Likelihood}
Following notation introduced in section (3.4) and Kumar~\etal~\cite{kumar2020luvli}, we formally define the Laplacian log-likelihood as:
\begin{align}
\mathcal{L}_{lll}(L^i_j, C^i_j, {L_{GT}}_j^i)_k = \frac{1}{2} log |\Sigma^i_{j,k}| + \sqrt{3(L^i_{j,k} - {L_{GT}}_{j,k}^i)^T(\Sigma^i_{j,k})^{-1}(L^i_{j,k} - {L_{GT}}_{j,k}^i)}
\end{align}
where, $\Sigma^i_{j,k}$ is the covariance matrix obtained from the Cholesky factor $C^i_{j,k}$ of the $k$th landmark of the $i$th FLSG of the $j$th dataset.

\section{300W Results}
We evaluate our method on the \TW~\cite{sagonas2013300} that contains 3,837 training images, and 600 testing images, with a 68 landmark definition. We train our model with two settings: \TW, and \TW~concurrently trained with \LAPA. We evaluate our model with inter-ocular normalization, and compare our results with state-of-the-art, Table~\ref{tbl:300w_comp}.
Here, we note that concurrent training with a larger dataset shows significant performance improvements.

\begin{table}
\center
\begin{tabular}{|l|l|l|l|}
\hline
Method    & \multicolumn{1}{c|}{\begin{tabular}[c]{@{}c@{}}Common\end{tabular}} & \multicolumn{1}{c|}{\begin{tabular}[c]{@{}c@{}}Challenge\end{tabular}} & \multicolumn{1}{c|}{Full} \\ \hline

PCD-CNN   & 3.67 & 7.62 & 4.44 \\ 
CPM+SBR   & 3.28 & 7.58 & 4.10 \\ 
SAN       & 3.34 & 6.60 & 3.98 \\ 
LAB       & 2.98 & 5.19 & 3.49 \\ 
DeCaFA    & 2.93 & 5.26 & 3.39 \\ 
U-Net    & 2.90 & 5.15 & 3.35 \\ 
HR-Net    & 2.85 & 5.15 & 3.32 \\ 
LUVLi     & 2.76 & 5.16 & 3.23 \\ 
AWing     & 2.72 & 4.52 & 3.07 \\ 
SH-FAN    & 2.61 & \textbf{4.13} & 2.94 \\ 
FaRL      & 2.56 & 4.45 & \textbf{2.93} \\ 
ADNet     & \textbf{2.53} & 4.58 & \textbf{2.93} \\ \hline 
MDMD Base & 2.91 & 5.12 & 3.34 \\ 
MDMD w/LaPa & 2.82 & 4.87 & 3.22 \\ \hline
\end{tabular}
\caption{Comparison against SOTA for \TW~\cite{sagonas2013300} on Inter-Ocular NME}
\label{tbl:300w_comp}
\vspace{-20px}
\end{table}

\section{Additional Implementation Details}

\subsection{Additional Architectural Details}\
\\
Our final prediction heads which regress the landmark and covariance information from the FLSG tokens each consist of two MLP heads. The covariance information is predicted by regressing the Cholesky factorization of the covariance matrix. Each MLP for landmarks and Cholesky prediction consist of two \texttt{relu} separated layers. The (input, output) dimensions for the first layer are \texttt{(768, 768//4)} for both head types and \texttt{(768//4, $N^i_j \times 2$)} and \texttt{(768//4, $N^i_j \times 3$)} for the second layer of the landmark and Cholesky heads respectively,  where $N_j^i$ is the number of landmarks for the $i$th FLSG and the $j$th dataset. 

\subsection{Augmentation Policy}\
\\
For training our model, we augment rigorously, applying random rotations, blurs, horizontal \& vertical waves, cutout, equalization, shear, color jitter, solarization, auto contrast, sharpness changes, posterization, inversion, scaling and translations, making use of~\cite{imgaug} for affine geometric transforms. We adopt two modified versions of Tan~\etal's ~\cite{Tan2019EfficientNetRM} AutoAugment~\cite{Cubuk2019AutoAugmentLA} policy, one which adds additional rotations and removes the translation, as we perform our translation augmentation later, and another which removes the geometric augmentations.

\subsection{FLSG Indexing Psuedocode Per (3.4)}
We present the pseudocode, as mentioned in section (3.4), for handling the FLSG heads and indexing:

\lstset{basicstyle=\tiny}
\begin{lstlisting}[language=Python]
class FLSGHead:
    def init(flsg_map: List[int]):
        flsg_map = flsg_map
        lm_heads = ModuleList(build_head(2*len(flsg)) for flsg in flsg_map)
        chol_heads = ModuleList(build_head(3*len(flsg)) for flsg in flsg_map)

    def build_head(flsg_dim: int):
        return Sequential(ReLU(), Linear(D, D // 4), ReLU(), Linear(D // 4, flsg_dim))

    def forward(flsg_tokens: Tensor):
        lms = concat([head(flsg_tokens[:, i]) for i, head in enum(lm_heads)])
        chols = concat([head(flsg_tokens[:, i]) for i, head in enum(chol_heads)])
        ids = [id for id_list in flsg_map for id in id_list] 
        return lms[:, ids], chols[:, ids]


class MDMDTransformer:
    def init():
        vit_encoder = ViT()
        flsg_maps = get_flsg_definitions() # [[lm_ids] * num_FLSGs] * num_datasets
        definition_agnostic_decoder = Decoder(flsg_maps)
        flsg_heads = ModuleList(FLSGHead(flsg_map) for flsg_map in flsg_maps)

    def forward(images: Tensor, dataset_id: int)
        image_features = vit_encoder(images)
        flsg_tokens = definition_agnostic_decoder(image_features)
        lms, chols = flsg_heads[dataset_id](flsg_tokens)
        return lms, chols
\end{lstlisting}

\subsubsection{FSLG Definitions}\
\\
We define the facial landmark semantic group definitions which were used for each dataset as follows:

\paragraph{Key:}
\begin{enumerate}[(a)]
    \item upper left contour
    \item lower left contour
    \item jaw
    \item lower right contour
    \item upper right contour
    \item left eye
    \item right eye
    \item left brow
    \item right brow
    \item nose
    \item top mouth
    \item bottom mouth
\end{enumerate}

\paragraph{Dataset Definitions}
\begin{enumerate}
    \item \WFLW~\cite{wayne2018lab}:
        \begin{enumerate}
            \item (0, 1, 2, 3, 4, 5)
            \item (6, 7, 8, 9, 10, 11, 12)
            \item (13, 14, 15, 16, 17, 18, 19)
            \item (20, 21, 22, 23, 24, 25, 26)
            \item (27, 28, 29, 30, 31, 32)
            \item (60, 61, 62, 63, 64, 65, 66, 67, 96)
            \item (68, 69, 70, 71, 72, 73, 74, 75, 97)
            \item (33, 34, 35, 36, 37, 38, 39, 40, 41)
            \item (42, 43, 44, 45, 46, 47, 48, 49, 50)
            \item (51, 52, 53, 54, 55, 56, 57, 58, 59)
            \item (77, 78, 79, 80, 81, 89, 90, 91)
            \item (76, 82, 83, 84, 85, 86, 87, 88, 92, 93, 94, 95)
        \end{enumerate}
    \item \LAPA~\cite{liu2019high}:
        \begin{enumerate}
            \item (0, 1, 2, 3, 4, 5)
            \item (6, 7, 8, 9, 10, 11, 12)
            \item (13, 14, 15, 16, 17, 18, 19)
            \item (20, 21, 22, 23, 24, 25, 26)
            \item (27, 28, 29, 30, 31, 32)
            \item (66, 67, 68, 69, 70, 71, 72, 73, 74, 104)
            \item (75, 76, 77, 78, 79, 80, 81, 82, 83, 105)
            \item (33, 34, 35, 36, 37, 38, 39, 40, 41)
            \item (42, 43, 44, 45, 46, 47, 48, 49, 50)
            \item (51, 52, 53, 54, 55, 56, 57, 58, 59, 60, 61, 62, 63, 64, 65)
            \item (85, 86, 87, 88, 89, 97, 98, 99)
            \item (84, 90, 91, 92, 93, 94, 95, 96, 100, 101, 102, 103)
        \end{enumerate}
    \item \COFW~\cite{burgos2013robust}:
        \begin{enumerate}
            \item -
            \item -
            \item (28)
            \item -
            \item -
            \item (8, 10, 12, 14, 16)
            \item (9, 11, 13, 15, 17)
            \item (0, 2, 4, 6)
            \item (1, 3, 5, 7)
            \item (18, 19, 20, 21)
            \item (22, 23, 24, 25)
            \item (26, 27)
        \end{enumerate}
    \item \TW~\cite{sagonas2013300}:
        \begin{enumerate}
            \item (0, 1, 2, 3)
            \item (4, 5, 6)
            \item (7, 8, 9)
            \item (10, 11, 1)
            \item (13, 14, 15, 16)
            \item (36, 37, 38, 39, 40, 41)
            \item (42, 43, 44, 45, 46, 47)
            \item (17, 18, 19, 20, 21)
            \item (22, 23, 24, 25, 26)
            \item (27, 28, 29, 30, 31, 32, 33, 34, 35)
            \item (48, 49, 50, 51, 52, 53, 54, 60, 61, 62, 63, 64)
            \item (55, 56, 57, 58, 59, 65, 66, 67)
        \end{enumerate}
    \item \AnimalWeb ~\cite{Khan2020AnimalWebAL}:
        \begin{enumerate}
            \item -
            \item -
            \item -
            \item -
            \item -
            \item (0, 1)
            \item (2, 3)
            \item -
            \item -
            \item (4)
            \item (5, 6, 7)
            \item (8)
        \end{enumerate}
    \item \ArtisticFaces ~\cite{Yaniv2019TheFO}:
        \begin{enumerate}
            \item (0, 1, 2, 3)
            \item (4, 5, 6)
            \item (7, 8, 9)
            \item (10, 11, 12)
            \item (13, 14, 15, 16)
            \item (36, 37, 38, 39, 40, 41)
            \item (42, 43, 44, 45, 46, 47)
            \item (17, 18, 19, 20, 21)
            \item (22, 23, 24, 25, 26)
            \item (27, 28, 29, 30, 31, 32, 33, 34, 35)
            \item (48, 49, 50, 51, 52, 53, 54, 60, 61, 62, 63, 64)
            \item (55, 56, 57, 58, 59, 65, 66, 67)
        \end{enumerate}
    \item \CaricatureFace ~\cite{Zhang2021LandmarkDA}:
        \begin{enumerate}
            \item (0, 1, 2, 3)
            \item (4, 5, 6)
            \item (7, 8, 9)
            \item (10, 11, 12)
            \item (13, 14, 15, 16)
            \item (36, 37, 38, 39, 40, 41)
            \item (42, 43, 44, 45, 46, 47)
            \item (17, 18, 19, 20, 21)
            \item (22, 23, 24, 25, 26)
            \item (27, 28, 29, 30, 31, 32, 33, 34, 35)
            \item (48, 49, 50, 51, 52, 53, 54, 60, 61, 62, 63, 64)
            \item (55, 56, 57, 58, 59, 65, 66, 67)
        \end{enumerate}
    \item \PARE~dataset [New]:
        \begin{enumerate}
            \item -
            \item -
            \item -
            \item -
            \item -
            \item (0, 1)
            \item (2, 3)
            \item -
            \item -
            \item (4)
            \item (5, 6, 7)
            \item (8)
        \end{enumerate}
\end{enumerate}

\section{PARE Dataset}
We release the labels for the PARE dataset containing 150 \emph{in-the-wild} illusory face images~\cite{wardle2022illusory} at the following: https://github.com/davidcferman/pareidolia-landmarks. The images and license information can be found at \url{https://osf.io/9g4rz/}.

\section{FLSG Groupings}
\begin{figure}[t]

  \includegraphics[width=\textwidth]{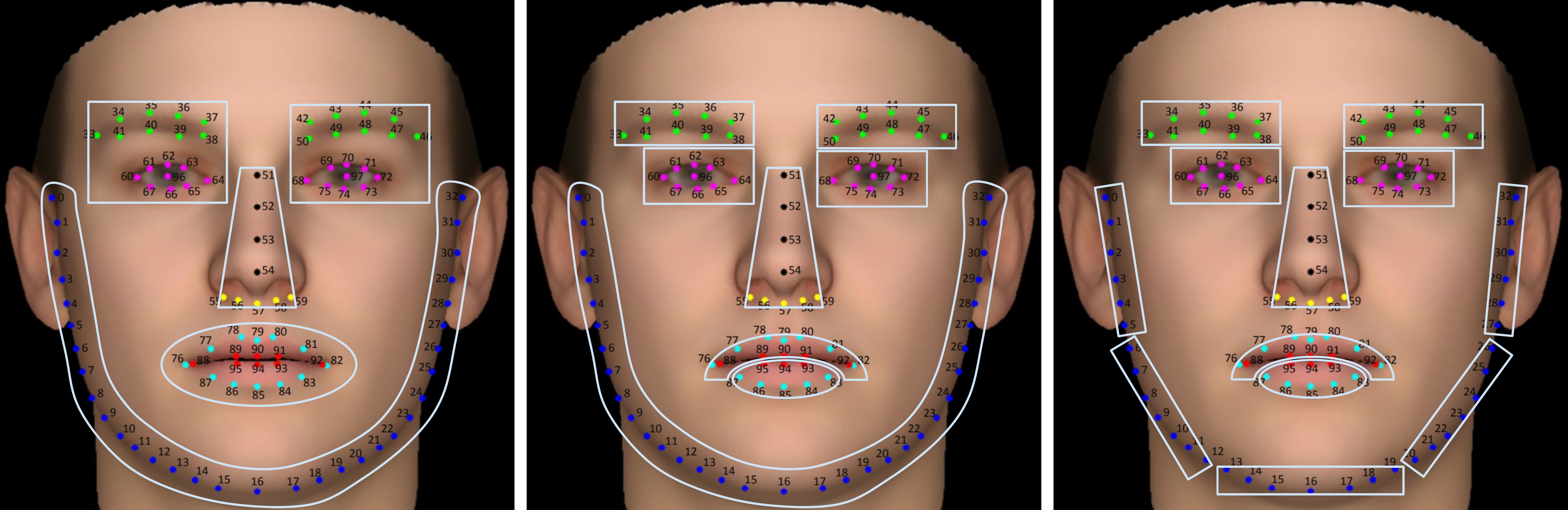}
  \caption{Facial Landmark Semantic Groupings. Image source: \cite{wayne2018lab}}
  \label{fig:flsg_groupings}
\end{figure}

We experiment with several FSLG grouping strategies, shown in Fig~\ref{fig:flsg_groupings}. The results from training on the \WFLW~\cite{wayne2018lab} dataset with each grouping strategy are shown in Table~\ref{tbl:flsg_comparison}. For our experiments, we selected the option with 12 FLSG groups, which performed best.

\begin{table}[t]
\center
\begin{tabular}{|l|l|l|l|}
\hline
Grouping    & \multicolumn{1}{c|}{NME$_{ic}$(\%)}  & \multicolumn{1}{c|}{FR$_{10\%}$}    & \multicolumn{1}{c|}{AUC$_{10\%}$} \\ \hline
5 Groups & \multicolumn{1}{c|}{4.12} & \multicolumn{1}{c|}{3.23} & \multicolumn{1}{c|}{59.43} \\
8 Groups & \multicolumn{1}{c|}{4.14} & \multicolumn{1}{c|}{2.88} & \multicolumn{1}{c|}{59.36} \\ 
12 Groups & \multicolumn{1}{c|}{\textbf{4.06}} & \multicolumn{1}{c|}{\textbf{2.63}} & \multicolumn{1}{c|}{\textbf{60.10}} \\ 
\hline
\end{tabular}
\caption{Comparison of FLSG grouping strategies on \WFLW~\cite{wayne2018lab}}
\label{tbl:flsg_comparison}
\end{table}

\section{\ArtisticFaces~\cite{Yaniv2019TheFO} Additional Comparisons}
We include additional comparisons against \ArtisticFaces. As previously mentioned, \ArtisticFaces's training set is a large set of style transferred images, while the testing set is 160 real paintings. However, our method trains on 112 of these real paintings, and tests on the remaining 48. We include comparisons when using the \ArtisticFaces~checkpoint on our 48 painting testing subset, for a direct comparison. Additionally, we include results with our method, trained on the style transferred images of \ArtisticFaces. We show the results in Table~\ref{tbl:artistic_faces_additional}.

\begin{table}[t]
\center
\begin{tabular}{|l|l|l|l|}
\hline
Method    & \multicolumn{1}{c|}{NME$_{ic}$(\%)} & Test Set \\ \hline
Yaniv et al.~\cite{Yaniv2019TheFO} & \multicolumn{1}{c|}{4.522} & Full Set \\
MDMD Base (style-transferred images) & \multicolumn{1}{c|}{3.996} & Full Set \\ \hline
Yaniv et al.~\cite{Yaniv2019TheFO} & \multicolumn{1}{c|}{4.573} & 30\% subset \\
MDMD Base & \multicolumn{1}{c|}{4.46} & 30\% subset \\
MDMD w/300W & \multicolumn{1}{c|}{3.72} & 30\% subset \\
\hline
\end{tabular}
\caption{Comparison against \ArtisticFaces~\cite{Yaniv2019TheFO}.}
\label{tbl:artistic_faces_additional}
\end{table}

\section{Backbone Comparisons}
We experiment with several backbone variations. While our model uses a pre-trained ViT backbone, we experiment with replacing this backbone with a Resnet-50, as well a Resnet-50 prior to our ViT. Additionally, we train our ViT from scratch for a similar number of epochs as we train our other models. We include results for ~\COFW~\cite{burgos2013robust} along with backbone parameter counts in Table~\ref{tbl:backbone_comparison}.

\begin{table}[t]
\center
\begin{tabular}{|l|l|l|l|l|}
\hline
Backbone    & \multicolumn{1}{c|}{NME$_{ip}$(\%)}  & \multicolumn{1}{c|}{FR$_{10\%}$}    & \multicolumn{1}{c|}{AUC$_{10\%}$} & Parameters \\ \hline
Resnet-50 & \multicolumn{1}{c|}{5.10} & \multicolumn{1}{c|}{.59} & \multicolumn{1}{c|}{49.12} & \multicolumn{1}{c|}{24 M} \\
Resnet-50 + ViT & \multicolumn{1}{c|}{5.72} & \multicolumn{1}{c|}{2.17} & \multicolumn{1}{c|}{42.92} & \multicolumn{1}{c|}{110 M} \\ 
ViT (scratch) & \multicolumn{1}{c|}{13.97} & \multicolumn{1}{c|}{60.2} & \multicolumn{1}{c|}{8.86} & \multicolumn{1}{c|}{86 M} \\ 
Early Convs~\cite{xiao2021early} + ViT & \multicolumn{1}{c|}{5.13} & \multicolumn{1}{c|}{1.18} & \multicolumn{1}{c|}{48.92} & \multicolumn{1}{c|}{86 M} \\ 
ViT & \multicolumn{1}{c|}{\textbf{4.82 }} & \multicolumn{1}{c|}{\textbf{.59}} & \multicolumn{1}{c|}{\textbf{51.84}} & \multicolumn{1}{c|}{86 M} \\ 
\hline
\end{tabular}
\caption{Comparison of various backbone strategies on~\COFW~\cite{burgos2013robust}.}
\label{tbl:backbone_comparison}
\end{table}

\section{Transfer Learning Comparison}
We compare our MDMD method to traditional transfer learning, both for \WFLW, trained with \LAPA, as well as \PARE, trained with \TW. Our model transfer learns from both the pre-trained backbone encoder and FLSG decoder. We include results in Table~\ref{tbl:tl_comparison}.

\begin{table}[t]
\center
\begin{tabular}{|l|l|l|l|}
\hline
Method    & \multicolumn{1}{c|}{NME$_{ic}$(\%)}  & \multicolumn{1}{c|}{FR$_{10\%}$}    & \multicolumn{1}{c|}{AUC$_{10\%}$} \\ \hline
MDMD \WFLW w/\LAPA & \multicolumn{1}{c|}{\textbf{3.97}} & \multicolumn{1}{c|}{2.2} & \multicolumn{1}{c|}{\textbf{.6083}} \\
TL~\LAPA~then \WFLW & \multicolumn{1}{c|}{4.00} & \multicolumn{1}{c|}{\textbf{1.94}} & \multicolumn{1}{c|}{.6074} \\ \hline
MDMD \PARE w/\TW & \multicolumn{1}{c|}{\textbf{8.59}} & \multicolumn{1}{c|}{\textbf{22.0}} & \multicolumn{1}{c|}{.2871} \\ 
TL \TW~then~\PARE & \multicolumn{1}{c|}{8.69} & \multicolumn{1}{c|}{24.0} & \multicolumn{1}{c|}{\textbf{.3004}} \\ 
\hline
\end{tabular}
\caption{Comparison of MDMD learning with traditional transfer learning (TL).}
\label{tbl:tl_comparison}
\end{table}

\end{document}